\pdfoutput=1
\documentclass{article}
\usepackage{ijcai23}
\usepackage{times}
\usepackage{soul}
\usepackage{url}
\usepackage[hidelinks]{hyperref}
\usepackage[utf8]{inputenc}
\usepackage[small]{caption}
\usepackage{graphicx}
\usepackage{amsmath}
\usepackage{amsthm}
\usepackage{booktabs}
\usepackage{algorithm}
\usepackage{algorithmic}
\usepackage[switch]{lineno}

\usepackage{xcolor}
\usepackage[flushleft]{threeparttable}
\usepackage{adjustbox}
\usepackage{multirow}
\usepackage{array}
\usepackage{pifont}
\newcommand{\cmark}{\ding{51}}%
\newcommand{\xmark}{\ding{55}}%

\usepackage{enumitem}
\newcommand{\methodname}[1]{\textbf{#1}} 

\usepackage{color, colortbl}
\definecolor{Gray}{gray}{0.89}

\newtheorem{problem}{Problem}

\title{Curriculum Graph Machine Learning: A Survey}

\author{
Haoyang Li\footnote{Equal contributions}
\and
Xin Wang\footnotemark[1] \and
Wenwu Zhu
\affiliations
Tsinghua University\\
\emails
lihy18@mails.tsinghua.edu.cn,
\{xin\_wang, wwzhu\}@tsinghua.edu.cn
}

\begin{document}

\maketitle

\begin{abstract}
Graph machine learning has been extensively studied in both academia and industry. However, in the literature, most existing graph machine learning models are designed to conduct training with data samples in a random order, which may suffer from suboptimal performance due to ignoring the importance of different graph data samples and their training orders for the model optimization status. To tackle this critical problem, curriculum graph machine learning (Graph CL), which integrates the strength of graph machine learning and curriculum learning, arises and attracts an increasing amount of attention from the research community. Therefore, in this paper, we comprehensively overview approaches on Graph CL and present a detailed survey of recent advances in this direction. Specifically, we first discuss the key challenges of Graph CL and provide its formal problem definition. Then, we categorize and summarize existing methods into three classes based on three kinds of graph machine learning tasks, i.e., node-level, link-level, and graph-level tasks. Finally, we share our thoughts on future research directions. To the best of our knowledge, this paper is the first survey for curriculum graph machine learning.
\end{abstract}

\section{Introduction}

Graph structured data is ubiquitous in the real world, which has been widely used to model the complex relationships and dependencies among various entities.
In the past decade, graph machine learning approaches, especially graph neural networks (GNNs)~\cite{kipf2016semi,velivckovic2017graph,hamilton2017inductive}, have drawn ever-increasing attention in both academia and industry, which make great progress in a variety of applications across wide-ranging domains, ranging from physics~\cite{sanchez2020learning} to chemistry~\cite{gilmer2017neural}, and from neuroscience~\cite{de2014graph} to social science~\cite{zhang2016final}.
Other areas, such as recommender systems~\cite{wu2022graph}, knowledge graphs~\cite{wang2017knowledge}, molecular prediction~\cite{hu2020open}, medical detection~\cite{horry2020covid}, drug repurposing~\cite{hsieh2021drug} etc., also provide an increasing demand for the applications of graph machine learning.

Despite the popularity of graph machine learning approaches, the existing literature generally trains graph models by feeding the data samples in a random order during the training process. 
For example, when training GNNs, the widely adopted mini-batch stochastic gradient descent optimization strategy as well as its variants select the data samples in each mini-batch randomly.
However, such training strategies largely ignore the importance of different graph data samples and how their orders can affect the optimization status, which may result in suboptimal performance of the graph learning models~\cite{wei2022clnode,wang2021curgraph}.
Typically, humans tend to learn much better when the data examples are organized in a meaningful order rather than randomly presented, e.g., learning from basic easy concepts to advanced hard concepts resembling the ``curriculum'' taught in schools~\cite{elman1993learning,rohde1999language}.
To this end, curriculum learning (CL)~\cite{bengio2009curriculum,survey1,survey2} is proposed to mimic human's learning process, 
and has been proved to be effective in boosting the model performances as well as improving the generalization capacity and convergence of learning models in various scenarios including computer vision~\cite{guo2018curriculumnet,jiang2014easy}, natural language processing~\cite{platanios2019competence,tay2019simple} etc.

Curriculum graph machine learning (Graph CL), which combines the strength of graph machine learning and curriculum learning, has become a promising research direction and attracted an increasing number of interests from the community, spanning over a variety of graph learning methodologies and applications recently.
Facing opportunities as well as challenges, we believe it is the right time to review and promote the studies of Graph CL approaches.
In this survey, we provide a comprehensive and systematic review of Graph CL. 
Specifically, we first analyze the key challenges of Graph CL and introduce the basic formulations. 
Then, we summarize the existing methodologies into three categories based on the granularity of graph tasks, i.e., node-level, link-level, and graph-level tasks, and elaborate representative approaches in each category.  
Last but not least, we discuss potential future research topics, which could shed light on the development of this promising area.
We deeply hope that this survey may provide insights for promoting the Graph CL research in the community.

\section{Challenges and Problem Formulation}\label{sec:challengeandformulation}

Curriculum graph machine learning (Graph CL), as an emerging research topic in the machine learning community, which non-trivially combines the strength of graph machine learning and curriculum  learning, faces the following challenges.
\begin{itemize}[leftmargin=*]\setlength{\itemsep}{0pt}
\item \textbf{Uniqueness of curriculum graph learning problem.} Unlike image or text, graph data lies in a non-Euclidean space. Besides, there exist complex relationships and dependencies between entities in graphs. Therefore, the non-Euclidean nature and complexity of graph-structured data bring unique challenges in tackling Graph CL problem. 
\item \textbf{Complexity of curriculum graph learning method.} Although general CL algorithms~\cite{survey1,survey2} have been extensively studied, it is non-trivial to combine the advantages of graph machine learning and curriculum learning into one unified framework, especially when customizing the design of the key CL components such as the difficulty measurer and the training scheduler that are compatible with graph models.
\item \textbf{Diversity of curriculum graph learning task.} Curriculum graph learning tasks range from node-level and link-level to graph-level problems with different settings, objectives, constraints, and domain knowledge. Therefore, it is important yet difficult to develop graph CL approaches tailored for different graph tasks or applications. 
\end{itemize}

The methodologies reviewed in this paper target on dealing with at least one of these three key challenges. 
Before introducing the problem formulation of curriculum graph machine learning (Graph CL), we briefly describe the background of graph machine learning and curriculum learning. 
 
\subsection{Graph Machine Learning}\label{sec:backgraphml}

Existing graph machine learning models can be generally divided into the following two groups: network embedding (NE)~\cite{cui2018survey,wang2017community} and graph neural networks (GNNs)~\cite{zhang2020deep}. Specifically, GNNs as the current state-of-the-art in graph machine learning, have been widely adopted to serve as the backbone of a graph curriculum learning method.

Let $\mathcal{G} = \left( \mathcal{V}, \mathcal{E}\right)$ be a graph with the node set $\mathcal{V}$ and the edge (link) set $\mathcal{E} \subseteq \mathcal{V} \times \mathcal{V}$. 
GNNs first learn node representations by the following message-passing~\cite{gilmer2017neural,zhang2022few} function:
\begin{equation}
\mathbf{h}_v^{(l+1)} = \mathrm{COM}\left (\mathbf{h}_v^{(l)}, \left[\mathrm{AGG}\left(\left\{\mathbf{h}_{u}^{(l)}~|~ u \in \mathcal{N}_v\right\}\right)\right]\right), 
\end{equation}
where $\mathbf{h}_v^{(l)}$ is the node representation of node $v$ at the $l^{th}$ layer, which is usually initialized as node feature at the first layer. $\mathcal{N}_v$ denotes the neighbors of node $v$. $\mathrm{AGG}(\cdot)$ and $\mathrm{COM}(\cdot)$ denote the aggregation and combination function of GNNs~\cite{wu2020comprehensive}.
After deriving the node representations, the link representations can be obtained based on the representations of the two connected nodes.
Furthermore, graph-level representations can be computed by the readout (or pooling)~\cite{xu2018powerful} operation on all nodes in this graph: 
$\mathbf{h}_G^{(l)} = \textrm{READOUT}\left\{\mathbf{h}_{v}^{(l)}~|~ v \in \mathcal{V}\right\}$. 
Finally, the representations of nodes, links, or graphs can be applied to different levels of graph tasks.

\subsection{Curriculum Learning}\label{sec:backcl}
Curriculum learning (CL), which mimics the human's learning process of learning data samples in a meaningful order, aims to enhance the machine learning models by using designed training curriculum, typically following an easy-to-hard pattern. As a general and flexible plug-in, the CL strategy has demonstrated its power in improving the model performance, generalization, robustness, and even convergence in a wide range of scenarios. In general, the framework of curriculum learning consists of a \emph{difficulty measurer} and a \emph{training scheduler}~\cite{survey1}. The difficulty measurer is adopted to calculate the difficulty score for each data sample and the training scheduler aims to arrange data samples in a meaningful order as the curriculum for training, based on the judgment of the difficulty measurer. 

The existing CL methods fall into two groups: (1) \emph{predefined CL} that manually designs heuristic-based policies to decide the training order, and (2) \emph{automatic CL} that relies on computable metrics (e.g., the training loss) to automatically design the curriculum for model training. 
Predefined CL can utilize expert knowledge but ignores the model's feedback, and automatic CL is domain-agnostic and more general as well as able to consider the model's feedback. That is to say, automatic CL is more data-driven or model-driven instead of human-driven, and more dynamically adaptive to the current training status, than predefined CL.
In addition, the training schedulers can be divided into \emph{discrete} and \emph{continuous} schedulers. The discrete schedulers adjust training after every fixed number ($> 1$) of epochs or convergence on the current data samples, while the continuous schedulers adjust the training at every epoch.

\subsection{Curriculum Graph Machine Learning}
Let $\mathcal{C} = \langle Q_1, \dots, Q_t, \dots, Q_T \rangle$ denotes a curriculum for training the graph model, which consists of a sequence of training data subsets over $T$ training steps. Each data subset $Q_t$ contains current training samples to be fed into the learning model at time step $t$. 
The order of all data subsets is determined by the difficulty measurer and the training scheduler.
Therefore, the problem of curriculum graph machine learning (Graph CL) can be formulated in the following:
\begin{problem} (Curriculum Graph Machine Learning). Given the training set of data instances (i.e., nodes, links, or graphs) $\mathcal{D} = \left\{(X_i,Y_i) \right\}_{i=1}^{|\mathcal{D}|}$, where $X_i$ is the input instance and $Y_i$ denotes the label, the goal is to optimize an optimal graph learning model 
guided by the curriculum $\mathcal{C}$, so as to achieve the best performance on testing graph data. 
\end{problem}

\begin{table*}[t]
\centering
\caption{A summary of curriculum graph machine learning methods. ``Graph CL Type'' denotes the curriculum learning type that each method belongs to, including predefined and automatic graph CL. ``Difficulty Measurer'' indicates the principle to design difficulty metric. ``Training Scheduler'' denotes the scheduler type that each method adopts. ``Task'' means the learning task of each method, including general tasks and specific applications. ``Need Label'' represents whether the method relies on labels during the training process.}
\label{table:properties}
\begin{adjustbox}{max width=1.0\textwidth}
\begin{tabular}{l|c|c|c|c|c}
\toprule
\textbf{Method}                         & \textbf{Graph CL Type} & \textbf{Difficulty Measurer} & \textbf{Training Scheduler} & \textbf{Task}               & \textbf{Need Label} \\
\midrule
\rowcolor{Gray}
\multicolumn{6}{c}{Node-level Graph CL}\\
\midrule
CLNode~\shortcite{wei2022clnode}        & Predefined             & Label Distribution           & Continuous                      & Node Classification         & \cmark              \\
GNN-CL~\shortcite{li2022graph}          & Predefined             & Sample Similarity            & Continuous                      & Node Classification         & \cmark              \\
SMMCL~\shortcite{gong2019multi}         & Predefined             & Label Distribution           & Discrete                        & Node Classification         & \cmark              \\
DiGCL~\shortcite{tong2021directed}      & Predefined             & Laplacian Perturbation      & Continuous                      & Node Classification         & \xmark              \\
HSAN~\shortcite{liu2022hard}            & Predefined             & Sample Similarity            & Discrete                        & Node Classification         & \xmark              \\
MTGNN~\shortcite{wu2020connecting}      & Predefined             & Step Length                  & Discrete                        & Time Series Forecasting      & \cmark              \\
MentorGNN~\shortcite{zhou2022mentorgnn} & Automatic              & Attention Weight             & Discrete                        & Node Classification         & \xmark              \\
RCL~\shortcite{anonymous2023relational} & Automatic              & Self-supervised Loss         & Continuous                      & Node Classification         & \cmark              \\
DRL~\shortcite{qu2018curriculum}        & Automatic              & Cumulative Reward            & Discrete                        & Node Classification         & \cmark              \\
GAUSS~\shortcite{guan2022large}         & Automatic              & Sample Loss                  & Discrete                        & Node Classification         & \cmark              \\
CGCT~\shortcite{roy2021curriculum}      & Automatic              & Sample Similarity            & Discrete                        & Image Classification        & \cmark              \\
\midrule
\rowcolor{Gray}
\multicolumn{6}{c}{Link-level Graph CL}\\
\midrule
GCN-WSRS~\shortcite{ying2018graph}      & Predefined             & Sample Similarity            & Continuous                      & Link Prediction             & \cmark              \\
TUNEUP~\shortcite{hu2022tuneup}         & Predefined             & Node Degree                  & Discrete                        & Link Prediction             & \cmark              \\
CHEST~\shortcite{wang2023curriculum}    & Predefined             & Pretraining Task             & Discrete                        & Link Prediction             & \xmark              \\
GTNN~\shortcite{vakil2022generic}       & Automatic              & Sample Loss                  & Discrete                        & Relation Extraction         & \cmark              \\
\midrule
\rowcolor{Gray}
\multicolumn{6}{c}{Graph-level Graph CL}\\
\midrule
CurGraph~\shortcite{wang2021curgraph}   & Predefined             & Label Distribution           & Discrete                        & Graph Classification        & \cmark              \\
CuCo~\shortcite{chu2021cuco}            & Predefined             & Sample Similarity            & Continuous                      & Graph Classification        & \xmark              \\
HACL~\shortcite{ahmed2022hyper}         & Predefined             & Sample Size                  & Discrete                        & Graph Classification        & \cmark              \\
Dual-GCN~\shortcite{dong2021dual}       & Predefined             & BLEU Metric                  & Discrete                        & Image Captioning            & \cmark              \\
CurrMG~\shortcite{gu2022curgraph}       & Automatic              & Domain Knowledge             & Continuous                      & Graph classification        & \cmark              \\
HAC-TSP~\shortcite{zhang2022learning}   & Automatic              & Solution Cost                & Continuous                      & Travelling Salesman Problem & \xmark              \\
\bottomrule
\end{tabular}
\end{adjustbox}
\end{table*}

\section{Categorization}

To tackle the non-trivial challenges introduced in Section~\ref{sec:challengeandformulation}, considerable efforts have been made in the literature, which integrate the strengths of graph machine learning and curriculum learning, and further propose tailored methods. 
Next, we will comprehensively review the existing methodologies by categorizing them into three groups based on the granularity of graph tasks, i.e., node-level, link-level, and graph-level tasks, followed by elaborations on representative approaches. Furthermore, for each category, we divide the methodologies into \emph{predefined graph CL} and \emph{automatic graph CL}, based on the type of curriculum learning strategy, i.e., whether adopting manually designed heuristic-based policies or automatically computable metrics to derive the curriculum for training.
The categorization and characteristics of surveyed graph CL methods are summarized in Table~\ref{table:properties}.

\section{Node-level Graph CL}\label{sec:nodecl}
Node is the fundamental unit for graph formation. Typically, the key to tackle node-level tasks is to learn node representations by training graph models (e.g., GNNs). Therefore, several studies on node-level graph CL are proposed to train graph models by starting with easy nodes and gradually include harder nodes during the training process, including both predefined and automatic node-level graph CL approaches.

\subsection{Predefined Node-level Graph CL}

Some methods define heuristic metrics to measure the difficulty of nodes in advance of the training process, e.g., by considering nodes' properties in terms of the topology, the feature, or the label.

\methodname{CLNode} (Curriculum Learning for Node Classification)~\cite{wei2022clnode} is a curriculum learning framework for node-level representation learning of GNNs. It boosts the performance of backbone GNNs via incrementally introducing nodes into the training process, starting with easy nodes and progressing to harder nodes.
The multi-perspective difficulty measurer is proposed to measure the difficulty of training nodes based on the label information. Specifically, the local perspective's difficulty measurer calculates local label distribution to recognize inter-class difficult nodes whose neighbors have diverse labels. And the global perspective's  difficulty measurer recognizes mislabeled difficult nodes in terms of node feature.
The continuous training scheduler is introduced to select appropriate training nodes in each epoch to mitigate the detrimental effect of difficult nodes.
CLNode can be compatible with most existing GNNs and boost their performances without increasing the time complexity just by feeding training nodes in order from easy to hard.

\methodname{GNN-CL} (Graph Neural Network with Curriculum Learning)~\cite{li2022graph} introduces curriculum learning into the imbalanced node classification task by controlling the training procedure from easy to hard, which consists of two modules.
The first one is an adaptive graph oversampling module, which interpolates the most significant samples related to the original structure, so as to dynamically adjust the data distribution in the graph from imbalance to balance.
The second one is a neighbor-based metric learning module. The distances between nodes and the connected neighbors are regularized based on the pseudo labels, which can dynamically adjust the position of the embeddings of minority class nodes in feature space. 
GNN-CL balances the label classification loss and neighbor-based triplet loss~\cite{wang2019dynamic} in the whole training process. 
The curriculum strategy consisting of two opposite learning curves is adopted. 
At the beginning of the training process, it pays more attention to optimizing feature propagation as well as reducing biased noises in the soft feature space. And it gradually focuses more on the average accuracy in each class, leading to strong accuracy on imbalanced node classification datasets.

\methodname{SMMCL} (Soft Multi-Modal Curriculum Learning)~\cite{gong2019multi} proposes a graph CL method for the label propagation on graphs~\cite{iscen2019label}. The goal is to learn the labels for predictions of unlabeled samples on graphs. 
Specifically, the authors assume that different unlabeled samples have different difficulty levels for propagation, so it should follow an easy-to-hard sequence with updated curriculum for label propagation.  
They also claim that the real-world graph data is often in multiple modalities~\cite{gao2017event}, where each modality should be equipped with a “teacher” that not only evaluates the difficulties of samples from its own viewpoint, but cooperates with other teachers to generate the overall simplest curriculum samples for propagation. 
They take the curriculum of the teachers as a whole, so that the common preference (i.e., commonality) of teachers on selecting samples can be captured.
Finally, an accurate curriculum sequence is established and the propagation quality can thus be improved, leading to more accurate label prediction results.

In addition to the supervised methods above that heavily rely on labels for training, there exist some self-supervised contrastive node-level graph CL methods.

\methodname{DiGCL} (Directed Graph Contrastive Learning)~\cite{tong2021directed} considers multi-task curriculum learning to progressively learn from multiple easy-to-hard contrastive views in directed graph contrastive learning. Specifically, to maximize the mutual information between the representations of different contrastive views~\cite{you2020graph} and produce informative node representations, it introduces a generalized dynamic-view contrastive objective.
The multi-task curriculum learning strategy is proposed to divide multiple contrastive views into sub-tasks with various difficulties and progressively learn from easy-to-hard sub-tasks. The different contrastive view pairs 
generated from Laplacian perturbation on the input graph
are scored by a difficulty measurer that is a predefined function in terms of the Laplacian perturbation~\cite{tong2020digraph}. And three common function families are considered as the training scheduler, including logarithmic, exponential, and linear functions. 
Note that this method is first to introduce curriculum learning in directed graph contrastive learning for node-level representation learning.

\methodname{HSAN} (Hard Sample Aware Network)~\cite{liu2022hard} also considers curriculum learning scheme in contrastive clustering on graphs. Specifically, it first introduces a similarity measure criterion between training pairs in graph contrastive learning, which measures the difficulty considering both attribute and structure information and improves the representativeness of the selected hard negative samples. Besides, to overcome the drawback of classical graph contrastive learning that the hard node pairs are treated equally, it proposes a dynamic weight modulating function to adjust the weights of sample pairs during training, which can up-weight the hard node pairs and down-weight the negative ones. The focusing factor controls the down-weighting rate of easy sample pairs, as the training scheduler of this method. Thus, the discriminative capability of learned representations is enhanced, leading to better performances.

\methodname{MTGNN} (Multivariate Time Series GNN)~\cite{wu2020connecting} is a graph CL method designed specifically for multivariate time series data~\cite{box2015time}, which advocates a curriculum learning strategy to find a better local optimum of the GNN and splits multivariate time series into subgroups during training. Specifically, since directly optimizing the traditional objective enforces the model to focus too much on improving the accuracy of long-term predictions while ignoring short-term ones, this method proposes a curriculum learning strategy for the multi-step forecasting task. The training is scheduled by starting with solving the easiest problem, i.e., only predicting the next one-step, which helps the model to find a good starting point. As the training progresses, the prediction length of the model is gradually increased, so that the model can learn the hard task step by step.
Overall, it is one effective trial combining GNN and curriculum learning in the application of multivariate time series forecasting.

\subsection{Automatic Node-level Graph CL}

We next describe automatic node-level graph CL methods, which consider the model's feedback during training to dynamically adapt to the optimization status.

\methodname{MentorGNN}~\cite{zhou2022mentorgnn} derives a curriculum for pre-training GNNs to learn informative node representations.
In order to tailor complex graph signals to boost the generalization performances, it develops a curriculum learning paradigm that automatically reweights graph signals for good generalization of the pre-trained GNNs in the target domain. Specifically, a teacher model that is a graph signal reweighting scheme gradually generates a domain-adaptive curriculum to guide the pre-training process of the student model that is a GNN architecture, so that the generalization performance in the node classification tasks can be enhanced. The curriculum is a sequence of graph signals that are extracted from the given graph. And the learned sample weighting scheme specifies a curriculum under which the GNNs are pre-trained gradually from the easy samples to the hard samples. The difficulty of training samples is measured by the teacher model and the training process is scheduled by the introduced learning threshold controlling the sample selection, so it belongs to automatic graph CL. The accuracy of this method on node classification in the graph transfer setting~\cite{hu2020pretraining} is largely enhanced.

\methodname{RCL} (Relational Curriculum Learning)~\cite{anonymous2023relational} claims that existing GNNs learn suboptimal node representations since they usually consider every edge of the input graph equally. 
Also, most graph CL methods simply consider nodes as independent samples for training, and introduce curriculum learning for these independent samples, which largely ignore the fundamental and unique dependency information behind the graph topology structure, and thus can not well deal with the correlation between nodes.
To tackle this problem, it proposes a graph CL method, which leverages the various underlying difficulties of data dependencies, to train better GNNs that can improve the quality of learned node representations. 
Specifically, it considers the relation between nodes gradually into training based on the relation's difficulty from easy to hard, where the degree of difficulty is measured by a self-supervised learning paradigm instead of a predefined heuristic-based metric. 
Then it develops an optimization model to iteratively increment the training structure according to the model training status and a theoretical guarantee of the convergence on the optimization algorithm is provided.
Finally, it presents an edge reweighting strategy to smooth the transition of the training structure between iterations, and reduce the influence of edges that connect nodes with relatively low confidence embeddings.

\methodname{DRL} (Learn curricula with Deep Reinforcement Learning)~\cite{qu2018curriculum} studies learning curriculum for node representations in heterogeneous star network~\cite{sun2009ranking} that has a center node type linked with multiple attribute node types through different types of edges, namely learning a sequence of edges of different types for node representation learning. 
A curriculum is defined as a sequence of edge types used for training, so the problem is formulated as a Markov decision process~\cite{puterman1990markov}.
It learns the optimal curriculum by estimating the $Q$ value of each state-action pair, namely the expected cumulative reward after taking the action from the state, by a planning module and a learning module.
Finally, the meaningful curriculum can be learned with high accuracy and low time costs for enhancing the performance of node classification.

\methodname{GAUSS} (Graph ArchitectUre Search at Scale)~\cite{guan2022large} is one large-scale node-level representation learning method by searching the GNN's architecture with curriculum learning. Since this method focuses on the large-scale graph whose efficiency issue becomes the main obstacle, it proposes a graph sampling-based single-path one-shot supernet~\cite{pham2018efficient} to reduce the computation burden.
To address the consistency collapse issues, it explicitly considers the joint architecture-graph sampling via a GNN architecture curriculum learning mechanism on the sampled sub-graphs and an architecture importance sampling algorithm~\cite{tokdar2010importance}. Specifically, it first forms a curriculum learning group for the GNN's architecture, and then makes the best learner as the teacher to decide a smoother learning objective for the group. The importance sampling is also utilized to reduce the variance during architecture sampling to form better learning group. Experiments of node classification on the large-scale graph show the effectiveness of this graph CL method.

\methodname{CGCT} (Curriculum Graph Co-Teaching)~\cite{roy2021curriculum} presents 
a graph CL method for the multi-target domain adaptation based on feature aggregation and curriculum learning.
The authors claim that learning robust and generalized representations in a unified space is just one prerequisite for tackling minimum risk across multiple target domains, where GNNs can play an important role in aggregating semantic information from neighbors across different domains on graphs represented as the source and target nodes. 
Then, the curriculum learning strategy is advocated in the proposed co-teaching framework to obtain pseudo-labels in an episodic fashion for mitigating information absence for the target nodes. 
Furthermore, an easy-to-hard curriculum learning strategy for domain selection is also proposed, where the feature alignment starts with the target domain that is closest to the source (easy one) and then gradually progresses towards the hard one, making the feature alignment process smoother. The experiments show that the performance on multi-target domain adaptation settings is improved by this curriculum graph co-teaching scheme.

\section{Link-level Graph CL}\label{sec:linkcl}

Links explicitly interconnect nodes in a graph, representing the relations and dependencies between nodes. Compared with node-level graph CL, link-level graph CL measures the difficulty and schedules the training on links. 

\subsection{Predefined Link-level Graph CL}

\methodname{GCN-WSRS} (GCN for Web-Scale Recommender Systems)~\cite{ying2018graph} presents a curriculum learning for negative sampling in link prediction tasks, whose effectiveness is validated in recommendation datasets.
Specifically, the GNN model for learning link representations is fed easy-to-hard samples during the training process, resulting in performance gains.
At the first training epoch, there are no hard negative links used for training, so the GNN model can quickly converge to find an area in the parameter space where the loss is small. 
Note that the hard samples are those related to the query samples, but not as related as the positive samples in link predictions.
The training scheduler outputs the rank of items in a graph according to the Personalized PageRank scores~\cite{gasteiger2018predict} with respect to the query sample.
As training proceeds, the harder negative links are added to the training set in subsequent epochs, which encourages the GNN model turns to learn how to
distinguish highly related samples from only slightly related ones. In particular, $e-1$ hard negative samples are added to the training set at the epoch $e$.

\methodname{TUNEUP}~\cite{hu2022tuneup} is a two-stage curriculum learning strategy for better training GNNs, showing competitive performance gains against traditional GNNs in predicting new links in a graph given existing links on citation networks, protein-protein networks, and recommendation benchmarks. The link scores are produced by the inner product of the two connected nodes.
Therefore, TUNEUP first trains a GNN to perform well on relatively easy head nodes (nodes with large degrees) and then proceeds to finetune the GNN to also perform well on hard tail nodes (nodes with small degrees). The node degrees are used to measure the difficulty of training samples. 
Specifically, in the first stage, TUNEUP randomly presents training samples to train the GNN model for obtaining a strong base GNN model. This base GNN model is encouraged to learn better representations for the head nodes, but performs poorly on the tail nodes. Thus, to handle this problem, in the second training stage, it further finetunes the base GNN model with increased supervisions on tail nodes, which follow the two steps: synthesizing more tail node inputs and adding target supervisions on the synthetic tail nodes.
In addition to improving the performance of link predictions, it also performs well in learning node representations.

\methodname{CHEST} (Curriculum pre-training based HEterogeneous Subgraph Transformer)~\cite{wang2023curriculum}  designs a curriculum graph pre-training strategy to gradually learn from both local and global contexts in the subgraph, which helps the GNN model to more effectively capture useful information for link prediction in recommendation. Compared with the supervised methods above, this method focuses on curriculum graph pre-training tailored to the link prediction. 
Specifically, the difficulty measurer is predefined on several pre-training tasks. Three easy pre-training curricula are related to node, edge and meta-path~\cite{sun2011pathsim}, focusing on local context information within the subgraph. And a hard pre-training curriculum is a subgraph contrastive learning task, focusing on global context information at subgraph level for user-item interaction.
The training scheduler is hand-crafted that schedules the pre-training tasks from the curricula in an ``easy-to-hard'' order, which is necessary to model complex data relations.
The learned GNN model after curriculum learning can produce representations that are aggregated into obvious clusters, by gradually extracting useful information for user-item interaction, to improve the link prediction tasks.

\subsection{Automatic Link-level Graph CL}

\methodname{GTNN} (Graph Text Neural Network)~\cite{vakil2022generic} is one representative automatic link-level graph CL method. 
It trains GNN with trend-aware curriculum learning, which improves the performance on relation extraction that is one type of link-level task. Inspired by the SuperLoss (SL)~\cite{castells2020superloss} which is a generic curriculum learning approach that dynamically learns a curriculum from model status, this method further designs Trend-SL curriculum learning approach, which belongs to self-paced curriculum learning~\cite{survey1}. Specifically, SuperLoss ignores the trend of instantaneous losses at sample-level that can not only improve the difficulty estimations of the model by making them local, sample dependent and potentially more precise, but also enable the model to distinguish samples with similar losses based on their known loss trajectories. In contrast, Trend-SL takes into account the loss information from the local time window before each iteration for capturing a form of momentum of loss on rising or falling trends and producing individual sample weights. Trend-SL adopts trend dynamics to shift the difficulty boundaries and adjust global difficulty using local sample-level loss trends.
With the help of the Trend-SL, the performance of relation extraction can be improved on several benchmarks.

\section{Graph-level Graph CL}\label{sec:graphcl}

Compared with node-level and link-level graph CL, graph-level graph CL focuses more on the global high-level properties of the whole graph to design the difficulty measurer and training scheduler for curriculum learning.

\subsection{Predefined Graph-level Graph CL}

\methodname{CurGraph}~\cite{wang2021curgraph} proposes a curriculum learning method for graph classification tasks via easy-to-hard curriculum. It first obtains graph-level embeddings via unsupervised GNN scheme InfoGraph~\cite{sun2019infograph} and further derives a neural density estimator to model embedding distributions. Then, to tackle the challenges of evaluating the difficulty of graphs induced by high irregular nature of graph data, it analyzes the graph difficulty in the high-level semantic feature space. The difficulty scores of graphs are calculated by a predefined difficulty measurer based on the intra-class and inter-class distributions of their embeddings. For the training scheduler, a smooth-step method is proposed to provide a soft transition from easy to hard graphs for GNNs. 
At each training step, the trained GNNs focus on the samples that are near the border of capability and neither too easy nor too hard, to expand the border gradually.
Finally, the performances of graph classification are enhanced without extra inference cost by feeding the graphs in an easy-to-hard order for training.

\methodname{CuCo}~\cite{chu2021cuco} incorporates curriculum learning into self-supervised graph-level representation learning. Compared with the supervised graph CL methods, this method focuses more on designing curriculum for selecting and training negative samples effectively, which are important steps in learning self-supervised graph representations. Similarly, it follows the learning process of humans by starting with easy negative samples when learning a new model and then learning difficult negative samples gradually. The difficulty measurer evaluates the difficulty of negative samples in the training dataset, which is calculated based on the embedding's similarity between the negative and positive samples. 
In addition, the training scheduler is proposed to decide how the negative samples are introduced to the training procedure by utilizing common function families: logarithmic, linear, quadratic, and root~\cite{survey1,survey2}.
This method is the first to study the impact of negative samples in graph-level contrastive representation learning by introducing curriculum learning.

\methodname{HACL} (Hyper-graph based Attention Curriculum Learning)~\cite{ahmed2022hyper} is a graph attention curriculum learning approach to learn hypergraph representations. Note that the hypergraphs are converted from the text written by the patients, which can be used to identify depressive symptoms of the patients.
Based on the semantic vectors from an emotion-driven context extraction technique and the structural hypergraph, this method separates the important boundary elements from the unlabeled sample and then incorporates them into the curriculum learning mechanism for the training process. 
Finally, the performances of classifying the nine particular symptoms are highly boosted due to the tailored designs on graph CL.

\methodname{Dual-GCN}~\cite{dong2021dual} is one GNN model encoding structure information from both local and global levels with curriculum learning as the training strategy, whose capability is improved on the image captioning task. 
Specifically, the local object-level GCN first converts one image into one graph, where the region to region relation in the image is modeled by the graph topology.  Besides, it also introduces auxiliary information to take into account the image’s similarity relation by the global image-level GCN.
To train this Dual-GCN, curriculum learning is adopted as the training strategy, where a cross-review scheme is introduced to distinguish the difficulty of the training samples.
The difficulty is measured by a predefined metric, which is used to sort the training samples and divide them into several subsets. After that, all the training samples are scheduled in the order from easy subsets to difficult subsets.

\subsection{Automatic Graph-level Graph CL}

\methodname{CurrMG}~\cite{gu2022curgraph} designs a curriculum learning strategy to learn representations for molecular graphs. 
 It arranges the easy-to-hard training curriculum in predefined, (automatic) transfer teacher, and hybrid types~\cite{hacohen2019power}. 
To design molecular graph CL with high robustness and applicability, six optional difficulty measurers, inspired by chemical domain knowledge and prior task information, are proposed, which can be divided into structure-specific, task-specific and hybrid types.
For the training scheduler, it is infeasible to split the training data based on molecular difficulties explicitly, as the difference of difficulty distributions is calculated by different difficulty measurers.
Therefore, a monotonically increasing smooth curve called competence function is introduced to control the data sampling space. 
The competence value is treated as a threshold for decreasing the data sampling space continuously, which is used for sampling current batch data during the training process.
Finally, the training scheduler encourages the GNN model to pay attention to the easy graphs in the early training stage, and gradually broaden the learning scope by contacting those difficult graphs.

\methodname{HAC-TSP} 
~\cite{zhang2022learning} proposes to design hardness-adaptive curriculum for solving travelling salesman problem (TSP)~\cite{applegate2011traveling} from the perspective of graphs. Compared with classical graph prediction tasks, defining a quantitative hardness measurement is challenging since obtaining the ground-truth solution of a TSP instance is NP-hard~\cite{hochba1997approximation}.  Therefore, the difficulty measurer calculates the hardness as greedy self-improving potentials by comparing the current cost with a surrogate model, avoiding the unbearable computational costs of calculating the ground-truth optimal solution for TSP. Then, a hardness-adaptive generator is designed to efficiently and continuously generate instances with different levels of hardness tailored for model training.
Finally, it proposes a curriculum learner to fully utilize the hardness-adaptive TSP instances. By learning instance weights, this method can train the TSP solvers more efficiently through curriculum learning. 
This method moves an important step towards graph CL for solving NP-hard problems.

\section{Future Directions}

Curriculum graph machine learning (Graph CL) is an emerging research topic.
Although significant progresses have been made for Graph CL, there still remain plenty of research directions worthy of future explorations.
\begin{itemize}[leftmargin=*]\setlength{\itemsep}{0pt}
\item \textbf{Theoretical guarantees}: Although various graph CL methods have been proposed and demonstrated effectiveness empirically, it remains to be further explored to derive more fundamental theoretical analysis on graph CL. A promising direction is to develop such theoretical analysis inspired by general curriculum learning from the perspective of optimization problem~\cite{weinshall2018curriculum} or data distribution~\cite{gong2016curriculum} for better understanding the mechanism and effectiveness of the graph CL methods.
\item \textbf{More principled Graph CL methods}: Although the existing works studied how curriculum learning methods are extended to graphs, it is still worth investigating to develop more principled models for graph curriculum learning by considering more detailed graph assumptions (e.g., homophily, heterophily), more complex graph types (e.g., attributed graphs, heterogeneous graphs, signed graphs, multiplex graphs), or more specific graph properties (e.g., data dependencies), into the model design. These methods are expected to further boost the model capacity.
\item \textbf{Generalization and transferability}: Currently, most existing graph CL methods are overly dependent on the graph labels, so that learned graph models tend to inherit a strong inductive bias for new testing tasks.  However, for real-world graphs, there will inevitably be scenarios with distribution shifts between testing and training graph data~\cite{li2022out,li2022ood}, which can induce significant performance drop for most existing approaches lacking the ability of generalization and transferability. One interesting direction is to learn label-irrelevant, generalizable and transferable representations via self-supervised learning via unsupervised pretext tasks, alleviating excessively relying on labeled data~\cite{chu2021cuco}. The other feasible direction is to explicitly consider distribution shifts in the design of graph CL methods for learning better graph representations~\cite{zhu2021shift}. 
\item \textbf{Comprehensive evaluation protocols}: While the graph CL methods have made great progress in performance boost, few works have made efforts on evaluating them with general graph benchmarks. The adopted datasets and evaluation metrics of existing works mainly follow classical settings. It is essential to develop a unified benchmark with unified metrics to evaluate and compare different methods, which should carefully incorporate datasets with different hardness (e.g., different levels of sparsity, heterogeneity, noises), and different evaluation metrics (e.g., relative performance boost, convergence speedup, additional computational costs). Besides, publicly available graph CL libraries are also important to facilitate and advance the research, which can be developed upon the existing library~\cite{curmllibrary}.
\item \textbf{Broader applications}: Although graph CL methods have been applied on several tasks as discussed in this work, it is worth further exploring the potential capacity in more diverse real-world applications, including recommendation~\cite{ying2018graph}, healthcare~\cite{ahmed2022hyper}, biochemistry~\cite{gu2022curgraph}, etc., for more effective and satisfying predictions. One of the major challenges is how to incorporate proper domain knowledge as additional priors to guide the model design. 
\end{itemize}

{
\clearpage
\newpage
\bibliographystyle{named}
\bibliography{ref}

\begin{thebibliography}{}

\bibitem[\protect\citeauthoryear{Ahmed \bgroup \em et al.\egroup }{2022}]{ahmed2022hyper}
Usman Ahmed, Jerry Chun-Wei Lin, and Gautam Srivastava.
\newblock Hyper-graph-based attention curriculum learning using a lexical algorithm for mental health.
\newblock {\em Pattern Recognition Letters}, 2022.

\bibitem[\protect\citeauthoryear{Applegate \bgroup \em et al.\egroup }{2011}]{applegate2011traveling}
David~L Applegate, Robert~E Bixby, Va{\v{s}}ek Chv{\'a}tal, and William~J Cook.
\newblock The traveling salesman problem.
\newblock In {\em The Traveling Salesman Problem}. Princeton university press, 2011.

\bibitem[\protect\citeauthoryear{Bengio \bgroup \em et al.\egroup }{2009}]{bengio2009curriculum}
Yoshua Bengio, J{\'e}r{\^o}me Louradour, Ronan Collobert, and Jason Weston.
\newblock Curriculum learning.
\newblock In {\em ICML}, 2009.

\bibitem[\protect\citeauthoryear{Box \bgroup \em et al.\egroup }{2015}]{box2015time}
George~EP Box, Gwilym~M Jenkins, Gregory~C Reinsel, and Greta~M Ljung.
\newblock {\em Time series analysis: forecasting and control}.
\newblock John Wiley \& Sons, 2015.

\bibitem[\protect\citeauthoryear{Castells \bgroup \em et al.\egroup }{2020}]{castells2020superloss}
Thibault Castells, Philippe Weinzaepfel, and Jerome Revaud.
\newblock Superloss: A generic loss for robust curriculum learning.
\newblock {\em NeurIPS}, 2020.

\bibitem[\protect\citeauthoryear{Chu \bgroup \em et al.\egroup }{2021}]{chu2021cuco}
Guanyi Chu, Xiao Wang, Chuan Shi, and Xunqiang Jiang.
\newblock Cuco: Graph representation with curriculum contrastive learning.
\newblock In {\em IJCAI}, 2021.

\bibitem[\protect\citeauthoryear{Cui \bgroup \em et al.\egroup }{2018}]{cui2018survey}
Peng Cui, Xiao Wang, Jian Pei, and Wenwu Zhu.
\newblock A survey on network embedding.
\newblock {\em IEEE TKDE}, 2018.

\bibitem[\protect\citeauthoryear{de Vico~Fallani \bgroup \em et al.\egroup }{2014}]{de2014graph}
Fabrizio de~Vico~Fallani, Jonas Richiardi, Mario Chavez, and Sophie Achard.
\newblock Graph analysis of functional brain networks: practical issues in translational neuroscience.
\newblock {\em Philosophical Transactions of the Royal Society B: Biological Sciences}, 2014.

\bibitem[\protect\citeauthoryear{Dong \bgroup \em et al.\egroup }{2021}]{dong2021dual}
Xinzhi Dong, Chengjiang Long, Wenju Xu, and Chunxia Xiao.
\newblock Dual graph convolutional networks with transformer and curriculum learning for image captioning.
\newblock In {\em ACM Multimedia}, 2021.

\bibitem[\protect\citeauthoryear{Elman}{1993}]{elman1993learning}
Jeffrey~L Elman.
\newblock Learning and development in neural networks: The importance of starting small.
\newblock {\em Cognition}, 1993.

\bibitem[\protect\citeauthoryear{Gao \bgroup \em et al.\egroup }{2017}]{gao2017event}
Yue Gao, Hanwang Zhang, Xibin Zhao, and Shuicheng Yan.
\newblock Event classification in microblogs via social tracking.
\newblock {\em ACM TIST}, 2017.

\bibitem[\protect\citeauthoryear{Gasteiger \bgroup \em et al.\egroup }{2018}]{gasteiger2018predict}
Johannes Gasteiger, Aleksandar Bojchevski, and Stephan G{\"u}nnemann.
\newblock Predict then propagate: Graph neural networks meet personalized pagerank.
\newblock {\em arXiv:1810.05997}, 2018.

\bibitem[\protect\citeauthoryear{Gilmer \bgroup \em et al.\egroup }{2017}]{gilmer2017neural}
Justin Gilmer, Samuel~S Schoenholz, Patrick~F Riley, Oriol Vinyals, and George~E Dahl.
\newblock Neural message passing for quantum chemistry.
\newblock In {\em ICML}, 2017.

\bibitem[\protect\citeauthoryear{Gong \bgroup \em et al.\egroup }{2016}]{gong2016curriculum}
Tieliang Gong, Qian Zhao, Deyu Meng, and Zongben Xu.
\newblock Why curriculum learning \& self-paced learning work in big/noisy data: A theoretical perspective.
\newblock {\em Big Data and Information Analytics}, 2016.

\bibitem[\protect\citeauthoryear{Gong \bgroup \em et al.\egroup }{2019}]{gong2019multi}
Chen Gong, Jian Yang, and Dacheng Tao.
\newblock Multi-modal curriculum learning over graphs.
\newblock {\em ACM TIST}, 2019.

\bibitem[\protect\citeauthoryear{Gu \bgroup \em et al.\egroup }{2022}]{gu2022curgraph}
Yaowen Gu, Si~Zheng, Zidu Xu, Qijin Yin, Liang Li, and Jiao Li.
\newblock {An efficient curriculum learning-based strategy for molecular graph learning}.
\newblock {\em Briefings in Bioinformatics}, 2022.

\bibitem[\protect\citeauthoryear{Guan \bgroup \em et al.\egroup }{2022}]{guan2022large}
Chaoyu Guan, Xin Wang, Hong Chen, Ziwei Zhang, and Wenwu Zhu.
\newblock Large-scale graph neural architecture search.
\newblock In {\em ICML}, 2022.

\bibitem[\protect\citeauthoryear{Guo \bgroup \em et al.\egroup }{2018}]{guo2018curriculumnet}
Sheng Guo, Weilin Huang, Haozhi Zhang, Chenfan Zhuang, Dengke Dong, Matthew~R Scott, and Dinglong Huang.
\newblock Curriculumnet: Weakly supervised learning from large-scale web images.
\newblock In {\em ECCV}, 2018.

\bibitem[\protect\citeauthoryear{Hacohen and Weinshall}{2019}]{hacohen2019power}
Guy Hacohen and Daphna Weinshall.
\newblock On the power of curriculum learning in training deep networks.
\newblock In {\em ICML}, 2019.

\bibitem[\protect\citeauthoryear{Hamilton \bgroup \em et al.\egroup }{2017}]{hamilton2017inductive}
Will Hamilton, Zhitao Ying, and Jure Leskovec.
\newblock Inductive representation learning on large graphs.
\newblock {\em NeurIPS}, 2017.

\bibitem[\protect\citeauthoryear{Hochba}{1997}]{hochba1997approximation}
Dorit~S Hochba.
\newblock Approximation algorithms for np-hard problems.
\newblock {\em ACM Sigact News}, 1997.

\bibitem[\protect\citeauthoryear{Horry \bgroup \em et al.\egroup }{2020}]{horry2020covid}
Michael~J Horry, Subrata Chakraborty, Manoranjan Paul, Anwaar Ulhaq, Biswajeet Pradhan, Manas Saha, and Nagesh Shukla.
\newblock Covid-19 detection through transfer learning using multimodal imaging data.
\newblock {\em Ieee Access}, 2020.

\bibitem[\protect\citeauthoryear{Hsieh \bgroup \em et al.\egroup }{2021}]{hsieh2021drug}
Kanglin Hsieh, Yinyin Wang, Luyao Chen, Zhongming Zhao, Sean Savitz, Xiaoqian Jiang, Jing Tang, and Yejin Kim.
\newblock Drug repurposing for covid-19 using graph neural network and harmonizing multiple evidence.
\newblock {\em Scientific reports}, 2021.

\bibitem[\protect\citeauthoryear{Hu \bgroup \em et al.\egroup }{2020a}]{hu2020open}
Weihua Hu, Matthias Fey, Marinka Zitnik, Yuxiao Dong, Hongyu Ren, Bowen Liu, Michele Catasta, and Jure Leskovec.
\newblock Open graph benchmark: Datasets for machine learning on graphs.
\newblock {\em NeurIPS}, 2020.

\bibitem[\protect\citeauthoryear{Hu \bgroup \em et al.\egroup }{2020b}]{hu2020pretraining}
Weihua Hu, Bowen Liu, Joseph Gomes, Marinka Zitnik, Percy Liang, Vijay Pande, and Jure Leskovec.
\newblock Strategies for pre-training graph neural networks.
\newblock In {\em ICLR}, 2020.

\bibitem[\protect\citeauthoryear{Hu \bgroup \em et al.\egroup }{2022}]{hu2022tuneup}
Weihua Hu, Kaidi Cao, Kexin Huang, Edward~W Huang, Karthik Subbian, and Jure Leskovec.
\newblock Tuneup: A training strategy for improving generalization of graph neural networks.
\newblock {\em arXiv:2210.14843}, 2022.

\bibitem[\protect\citeauthoryear{Iscen \bgroup \em et al.\egroup }{2019}]{iscen2019label}
Ahmet Iscen, Giorgos Tolias, Yannis Avrithis, and Ondrej Chum.
\newblock Label propagation for deep semi-supervised learning.
\newblock In {\em CVPR}, 2019.

\bibitem[\protect\citeauthoryear{Jiang \bgroup \em et al.\egroup }{2014}]{jiang2014easy}
Lu~Jiang, Deyu Meng, Teruko Mitamura, and Alexander~G Hauptmann.
\newblock Easy samples first: Self-paced reranking for zero-example multimedia search.
\newblock In {\em ACM Multimedia}, 2014.

\bibitem[\protect\citeauthoryear{Kipf and Welling}{2017}]{kipf2016semi}
Thomas~N Kipf and Max Welling.
\newblock Semi-supervised classification with graph convolutional networks.
\newblock In {\em ICLR}, 2017.

\bibitem[\protect\citeauthoryear{Li \bgroup \em et al.\egroup }{2022a}]{li2022ood}
Haoyang Li, Xin Wang, Ziwei Zhang, and Wenwu Zhu.
\newblock Ood-gnn: Out-of-distribution generalized graph neural network.
\newblock {\em IEEE TKDE}, 2022.

\bibitem[\protect\citeauthoryear{Li \bgroup \em et al.\egroup }{2022b}]{li2022out}
Haoyang Li, Xin Wang, Ziwei Zhang, and Wenwu Zhu.
\newblock Out-of-distribution generalization on graphs: A survey.
\newblock {\em arXiv:2202.07987}, 2022.

\bibitem[\protect\citeauthoryear{Li \bgroup \em et al.\egroup }{2022c}]{li2022graph}
Xiaohe Li, Lijie Wen, Yawen Deng, Fuli Feng, Xuming Hu, Lei Wang, and Zide Fan.
\newblock Graph neural network with curriculum learning for imbalanced node classification.
\newblock {\em arXiv:2202.02529}, 2022.

\bibitem[\protect\citeauthoryear{Liu \bgroup \em et al.\egroup }{2023}]{liu2022hard}
Yue Liu, Xihong Yang, Sihang Zhou, Xinwang Liu, Zhen Wang, Ke~Liang, Wenxuan Tu, Liang Li, Jingcan Duan, and Cancan Chen.
\newblock Hard sample aware network for contrastive deep graph clustering.
\newblock {\em AAAI}, 2023.

\bibitem[\protect\citeauthoryear{Pham \bgroup \em et al.\egroup }{2018}]{pham2018efficient}
Hieu Pham, Melody Guan, Barret Zoph, Quoc Le, and Jeff Dean.
\newblock Efficient neural architecture search via parameters sharing.
\newblock In {\em ICML}, 2018.

\bibitem[\protect\citeauthoryear{Platanios \bgroup \em et al.\egroup }{2019}]{platanios2019competence}
Emmanouil~Antonios Platanios, Otilia Stretcu, Graham Neubig, Barnabas Poczos, and Tom~M Mitchell.
\newblock Competence-based curriculum learning for neural machine translation.
\newblock {\em arXiv:1903.09848}, 2019.

\bibitem[\protect\citeauthoryear{Puterman}{1990}]{puterman1990markov}
Martin~L Puterman.
\newblock Markov decision processes.
\newblock {\em Handbooks in operations research and management science}, 1990.

\bibitem[\protect\citeauthoryear{Qu \bgroup \em et al.\egroup }{2018}]{qu2018curriculum}
Meng Qu, Jian Tang, and Jiawei Han.
\newblock Curriculum learning for heterogeneous star network embedding via deep reinforcement learning.
\newblock In {\em WSDM}, 2018.

\bibitem[\protect\citeauthoryear{Rohde and Plaut}{1999}]{rohde1999language}
Douglas~LT Rohde and David~C Plaut.
\newblock Language acquisition in the absence of explicit negative evidence: How important is starting small?
\newblock {\em Cognition}, 1999.

\bibitem[\protect\citeauthoryear{Roy \bgroup \em et al.\egroup }{2021}]{roy2021curriculum}
Subhankar Roy, Evgeny Krivosheev, Zhun Zhong, Nicu Sebe, and Elisa Ricci.
\newblock Curriculum graph co-teaching for multi-target domain adaptation.
\newblock In {\em CVPR}, 2021.

\bibitem[\protect\citeauthoryear{Sanchez-Gonzalez \bgroup \em et al.\egroup }{2020}]{sanchez2020learning}
Alvaro Sanchez-Gonzalez, Jonathan Godwin, Tobias Pfaff, Rex Ying, Jure Leskovec, and Peter Battaglia.
\newblock Learning to simulate complex physics with graph networks.
\newblock In {\em ICML}, 2020.

\bibitem[\protect\citeauthoryear{Soviany \bgroup \em et al.\egroup }{2022}]{survey2}
Petru Soviany, Radu~Tudor Ionescu, Paolo Rota, and Nicu Sebe.
\newblock Curriculum learning: A survey.
\newblock {\em IJCV}, 2022.

\bibitem[\protect\citeauthoryear{Sun \bgroup \em et al.\egroup }{2009}]{sun2009ranking}
Yizhou Sun, Yintao Yu, and Jiawei Han.
\newblock Ranking-based clustering of heterogeneous information networks with star network schema.
\newblock In {\em SIGKDD}, 2009.

\bibitem[\protect\citeauthoryear{Sun \bgroup \em et al.\egroup }{2011}]{sun2011pathsim}
Yizhou Sun, Jiawei Han, Xifeng Yan, Philip~S Yu, and Tianyi Wu.
\newblock Pathsim: Meta path-based top-k similarity search in heterogeneous information networks.
\newblock {\em VLDB Endowment}, 2011.

\bibitem[\protect\citeauthoryear{Sun \bgroup \em et al.\egroup }{2019}]{sun2019infograph}
Fan-Yun Sun, Jordan Hoffman, Vikas Verma, and Jian Tang.
\newblock Infograph: Unsupervised and semi-supervised graph-level representation learning via mutual information maximization.
\newblock In {\em ICLR}, 2019.

\bibitem[\protect\citeauthoryear{Tay \bgroup \em et al.\egroup }{2019}]{tay2019simple}
Yi~Tay, Shuohang Wang, Luu~Anh Tuan, Jie Fu, Minh~C Phan, Xingdi Yuan, Jinfeng Rao, Siu~Cheung Hui, and Aston Zhang.
\newblock Simple and effective curriculum pointer-generator networks for reading comprehension over long narratives.
\newblock {\em arXiv:1905.10847}, 2019.

\bibitem[\protect\citeauthoryear{Tokdar and Kass}{2010}]{tokdar2010importance}
Surya~T Tokdar and Robert~E Kass.
\newblock Importance sampling: a review.
\newblock {\em Wiley Interdisciplinary Reviews: Computational Statistics}, 2010.

\bibitem[\protect\citeauthoryear{Tong \bgroup \em et al.\egroup }{2020}]{tong2020digraph}
Zekun Tong, Yuxuan Liang, Changsheng Sun, Xinke Li, David Rosenblum, and Andrew Lim.
\newblock Digraph inception convolutional networks.
\newblock {\em NeurIPS}, 2020.

\bibitem[\protect\citeauthoryear{Tong \bgroup \em et al.\egroup }{2021}]{tong2021directed}
Zekun Tong, Yuxuan Liang, Henghui Ding, Yongxing Dai, Xinke Li, and Changhu Wang.
\newblock Directed graph contrastive learning.
\newblock {\em NeurIPS}, 2021.

\bibitem[\protect\citeauthoryear{Vakil and Amiri}{2022}]{vakil2022generic}
Nidhi Vakil and Hadi Amiri.
\newblock Generic and trend-aware curriculum learning for relation extraction in graph neural networks.
\newblock {\em arXiv:2205.08625}, 2022.

\bibitem[\protect\citeauthoryear{Veli{\v{c}}kovi{\'c} \bgroup \em et al.\egroup }{2018}]{velivckovic2017graph}
Petar Veli{\v{c}}kovi{\'c}, Guillem Cucurull, Arantxa Casanova, Adriana Romero, Pietro Lio, and Yoshua Bengio.
\newblock Graph attention networks.
\newblock In {\em ICLR}, 2018.

\bibitem[\protect\citeauthoryear{Wang \bgroup \em et al.\egroup }{2017a}]{wang2017knowledge}
Quan Wang, Zhendong Mao, Bin Wang, and Li~Guo.
\newblock Knowledge graph embedding: A survey of approaches and applications.
\newblock {\em IEEE TKDE}, 2017.

\bibitem[\protect\citeauthoryear{Wang \bgroup \em et al.\egroup }{2017b}]{wang2017community}
Xiao Wang, Peng Cui, Jing Wang, Jian Pei, Wenwu Zhu, and Shiqiang Yang.
\newblock Community preserving network embedding.
\newblock In {\em AAAI}, 2017.

\bibitem[\protect\citeauthoryear{Wang \bgroup \em et al.\egroup }{2019}]{wang2019dynamic}
Yiru Wang, Weihao Gan, Jie Yang, Wei Wu, and Junjie Yan.
\newblock Dynamic curriculum learning for imbalanced data classification.
\newblock In {\em ICCV}, 2019.

\bibitem[\protect\citeauthoryear{Wang \bgroup \em et al.\egroup }{2021a}]{survey1}
Xin Wang, Yudong Chen, and Wenwu Zhu.
\newblock A survey on curriculum learning.
\newblock {\em IEEE TPAMI}, 2021.

\bibitem[\protect\citeauthoryear{Wang \bgroup \em et al.\egroup }{2021b}]{wang2021curgraph}
Yiwei Wang, Wei Wang, Yuxuan Liang, Yujun Cai, and Bryan Hooi.
\newblock Curgraph: Curriculum learning for graph classification.
\newblock In {\em The WebConf}, 2021.

\bibitem[\protect\citeauthoryear{Wang \bgroup \em et al.\egroup }{2023}]{wang2023curriculum}
Hui Wang, Kun Zhou, Xin Zhao, Jingyuan Wang, and Ji-Rong Wen.
\newblock Curriculum pre-training heterogeneous subgraph transformer for top-n recommendation.
\newblock {\em TOIS}, 2023.

\bibitem[\protect\citeauthoryear{Wei \bgroup \em et al.\egroup }{2022}]{wei2022clnode}
Xiaowen Wei, Weiwei Liu, Yibing Zhan, Du~Bo, and Wenbin Hu.
\newblock Clnode: Curriculum learning for node classification.
\newblock {\em arXiv:2206.07258}, 2022.

\bibitem[\protect\citeauthoryear{Weinshall \bgroup \em et al.\egroup }{2018}]{weinshall2018curriculum}
Daphna Weinshall, Gad Cohen, and Dan Amir.
\newblock Curriculum learning by transfer learning: Theory and experiments with deep networks.
\newblock In {\em ICML}, 2018.

\bibitem[\protect\citeauthoryear{Wu \bgroup \em et al.\egroup }{2020a}]{wu2020comprehensive}
Zonghan Wu, Shirui Pan, Fengwen Chen, Guodong Long, Chengqi Zhang, and S~Yu Philip.
\newblock A comprehensive survey on graph neural networks.
\newblock {\em IEEE TNNLS}, 2020.

\bibitem[\protect\citeauthoryear{Wu \bgroup \em et al.\egroup }{2020b}]{wu2020connecting}
Zonghan Wu, Shirui Pan, Guodong Long, Jing Jiang, Xiaojun Chang, and Chengqi Zhang.
\newblock Connecting the dots: Multivariate time series forecasting with graph neural networks.
\newblock In {\em SIGKDD}, 2020.

\bibitem[\protect\citeauthoryear{Wu \bgroup \em et al.\egroup }{2022}]{wu2022graph}
Shiwen Wu, Fei Sun, Wentao Zhang, Xu~Xie, and Bin Cui.
\newblock Graph neural networks in recommender systems: a survey.
\newblock {\em ACM Computing Surveys}, 2022.

\bibitem[\protect\citeauthoryear{Xu \bgroup \em et al.\egroup }{2019}]{xu2018powerful}
Keyulu Xu, Weihua Hu, Jure Leskovec, and Stefanie Jegelka.
\newblock How powerful are graph neural networks?
\newblock {\em ICLR}, 2019.

\bibitem[\protect\citeauthoryear{Ying \bgroup \em et al.\egroup }{2018}]{ying2018graph}
Rex Ying, Ruining He, Kaifeng Chen, Pong Eksombatchai, William~L Hamilton, and Jure Leskovec.
\newblock Graph convolutional neural networks for web-scale recommender systems.
\newblock In {\em SIGKDD}, 2018.

\bibitem[\protect\citeauthoryear{You \bgroup \em et al.\egroup }{2020}]{you2020graph}
Yuning You, Tianlong Chen, Yongduo Sui, Ting Chen, Zhangyang Wang, and Yang Shen.
\newblock Graph contrastive learning with augmentations.
\newblock {\em NeurIPS}, 2020.

\bibitem[\protect\citeauthoryear{Zhang and Tong}{2016}]{zhang2016final}
Si~Zhang and Hanghang Tong.
\newblock Final: Fast attributed network alignment.
\newblock In {\em SIGKDD}, 2016.

\bibitem[\protect\citeauthoryear{Zhang \bgroup \em et al.\egroup }{2020}]{zhang2020deep}
Ziwei Zhang, Peng Cui, and Wenwu Zhu.
\newblock Deep learning on graphs: A survey.
\newblock {\em IEEE TKDE}, 2020.

\bibitem[\protect\citeauthoryear{Zhang \bgroup \em et al.\egroup }{2022a}]{zhang2022few}
Chuxu Zhang, Kaize Ding, Jundong Li, Xiangliang Zhang, Yanfang Ye, Nitesh~V Chawla, and Huan Liu.
\newblock Few-shot learning on graphs: A survey.
\newblock {\em IJCAI}, 2022.

\bibitem[\protect\citeauthoryear{Zhang \bgroup \em et al.\egroup }{2022b}]{zhang2022learning}
Zeyang Zhang, Ziwei Zhang, Xin Wang, and Wenwu Zhu.
\newblock Learning to solve travelling salesman problem with hardness-adaptive curriculum.
\newblock In {\em AAAI}, 2022.

\bibitem[\protect\citeauthoryear{Zhang \bgroup \em et al.\egroup }{2023}]{anonymous2023relational}
Zheng Zhang, Junxiang Wang, and Liang Zhao.
\newblock Curriculum learning for graph neural networks: Which edges should we learn first.
\newblock {\em NeurIPS}, 2023.

\bibitem[\protect\citeauthoryear{Zhou \bgroup \em et al.\egroup }{2022a}]{zhou2022mentorgnn}
Dawei Zhou, Lecheng Zheng, Dongqi Fu, Jiawei Han, and Jingrui He.
\newblock Mentorgnn: deriving curriculum for pre-training gnns.
\newblock In {\em CIKM}, 2022.

\bibitem[\protect\citeauthoryear{Zhou \bgroup \em et al.\egroup }{2022b}]{curmllibrary}
Yuwei Zhou, Hong Chen, Zirui Pan, Chuanhao Yan, Fanqi Lin, Xin Wang, and Wenwu Zhu.
\newblock Curml: A curriculum machine learning library.
\newblock In {\em ACM Multimedia}, 2022.

\bibitem[\protect\citeauthoryear{Zhu \bgroup \em et al.\egroup }{2021}]{zhu2021shift}
Qi~Zhu, Natalia Ponomareva, Jiawei Han, and Bryan Perozzi.
\newblock Shift-robust gnns: Overcoming the limitations of localized graph training data.
\newblock {\em NeurIPS}, 2021.

\end{thebibliography}
}
\end{document}